\documentclass{llncs}
\usepackage{enumerate}
\usepackage{amsfonts}
\usepackage{amsmath}
\usepackage{color}
\usepackage{tikz}
\usepackage{url} 
\usepackage[procnames]{listings}
\usepackage{color}
\usepackage{todonotes}
\usepackage{colortbl}
\usepackage{pgfplots} 
\usepackage{times} 

\def\tensorflow{\textsc{TensorFlow}$^{\text{\tiny TM}}$}
\newcommand{\argmin}{\mathop{\mathrm{argmin}}}

\definecolor{keywords}{RGB}{255,0,90}
\definecolor{comments}{RGB}{0,0,113}
\definecolor{red}{RGB}{160,0,0}
\definecolor{green}{RGB}{0,150,0}
 
\title{Logic Tensor Networks: Deep Learning and Logical Reasoning from Data and Knowledge%
\thanks{The first author acknowledges the Mobility Program of FBK, for
  supporting a long term visit at City University London. He also acknowledges
  NVIDIA Corporation for supporting this research with the donation of
  a GPU.}} 
\author{Luciano Serafini \inst{1} \and Artur d'Avila Garcez \inst{2}}
\institute{Fondazione Bruno Kessler, Trento, Italy, \texttt{serafini@fbk.eu} \and 
                      City University London, UK, \texttt{a.garcez@city.ac.uk}}

\def\BOmega{\mathbf{\Omega}}
\def\K{\mathcal{K}}
\def\L{\mathcal{L}}
\def\G{\mathcal{G}}
\def\C{\mathcal{C}}
\def\Fn{\mathcal{F}}
\def\Pr{\mathcal{P}}
\def\R{\mathbb{R}}

\def\imp{\rightarrow}
\def\bx{\mathbf{x}}

\def\bv{\mathbf{v}}

\def\ltn{\texttt{ltn}}
\begin{document}
\maketitle

\begin{abstract}
We propose \emph{Logic Tensor Networks}: a uniform framework for
integrating automatic learning and reasoning. A logic formalism called 
Real Logic is defined on a first-order language whereby formulas 
have truth-value in the interval [0,1] and semantics defined 
concretely on the domain of real numbers. Logical constants are
interpreted as feature vectors of real numbers. Real Logic 
promotes a well-founded integration of deductive reasoning on 
a knowledge-base and efficient data-driven relational machine 
learning. We show how Real Logic can be implemented in 
deep Tensor Neural Networks with the use of Google's \tensorflow\ 
primitives. The paper concludes with experiments applying 
Logic Tensor Networks on a simple but 
representative example of knowledge completion. 
\end{abstract}

 
\begin{description}
\item[Keywords:]
  Knowledge Representation, Relational Learning, Tensor Networks, 
  Neural-Symbolic Computation, Data-driven Knowledge Completion.
\end{description}
\def\arity{\alpha}
\def\Rel{\mathcal{R}}
\def\Citation{\mathit{Cite}}
\def\MoreRecent{\mathit{MoreRecent}}
\def\Influential{\mathit{Influential}}
\def\Seminal{\mathit{Seminal}}
\def\Similar{\mathit{Similar}}

\section{Introduction}

The recent availability of large-scale data combining multiple data
modalities, such as image, text, audio and sensor data, has opened up
various research and commercial opportunities, underpinned by machine
learning methods and techniques \cite{Bengio:2009:LDA:1658423.1658424,44806,Kephart:2003:VAC:642194.642200,kiela-bottou-2014}. In particular, recent work in
machine learning has sought to combine logical services, such as
knowledge completion, approximate inference, and goal-directed
reasoning with data-driven statistical and neural network-based
approaches.  We argue that there are great possibilities for improving
the current state of the art in machine learning and artificial
intelligence (AI) thought the principled combination of knowledge
representation, reasoning and learning.  Guha's recent position paper
\cite{towards-a-model-theory-for-distributed-representations-guha2015} is a case
in point, as it advocates a new model theory for real-valued numbers.
In this paper, we take inspiration from such recent work in AI, but also less recent 
work in the area of neural-symbolic integration \cite{tr-bottou-2011,DBLP:series/cogtech/GarcezLG2009,DBLP:journals/ml/DiligentiGMR12} and in semantic attachment and symbol grounding \cite{DBLP:journals/neco/BarrettFD08} 
to achieve a vector-based representation which can be shown adequate for integrating 
machine learning and reasoning in a principled way. 

This paper proposes a framework called \emph{Logic Tensor Networks (LTN)} which 
integrates learning based on tensor networks \cite{SocherChenManningNg2013} 
with reasoning using first-order many-valued logic \cite{bergmann2008introduction}, all implemented in \tensorflow\ \cite{tensorflow2015-whitepaper}. This enables, for the first 
time, a range of knowledge-based tasks using rich knowledge representation in 
first-order logic (FOL) to be combined with efficient data-driven machine learning based on 
the manipulation of real-valued vectors\footnote{In practice, FOL reasoning including function symbols is approximated through the usual iterative deepening of clause depth.}. 
Given data available in the form of 
real-valued vectors, logical soft and hard constraints and relations which apply to
certain subsets of the vectors can be specified compactly in first-order logic. 
Reasoning about such constraints can help improve learning, and learning from new data 
can revise such constraints thus modifying reasoning. An adequate vector-based 
representation of the logic, first proposed in this paper, enables the above integration 
of learning and reasoning, as detailed in what follows.

We are interested in providing a computationally adequate approach
to implementing learning and reasoning \cite{Valiant:1999:RL:301250.301425} in an integrated way within an idealized
agent. This agent has to manage knowledge about 
an unbounded, possibly infinite, set of objects $O=\{o_1,o_2,\dots\}$. 
Some of the objects are associated with
a set of quantitative attributes, represented by
an $n$-tuple of real values $\G(o_i)\in\R^n$, which we call \emph{grounding}. 
For example, a person may have a grounding into a $4$-tuple containing 
some numerical representation of the person's name, her height,
weight, and number of friends in some social network. 
Object tuples can participate in 
a set of relations $\Rel=\{R_1,\dots,R_k\}$, with $R_i\subseteq O^{\arity(R_i)}$, 
where $\arity(R_i)$ denotes the arity of relation $R_i$. We presuppose the 
existence of a latent (unknown) relation between the above numerical properties, i.e. 
groundings, and partial relational structure $\Rel$ on $O$.  
Starting from this partial knowledge, an agent is required to: (i) 
infer new knowledge about the relational structure on the
  objects of $O$; (ii) predict the numerical properties or the class of the objects in $O$.

Classes and relations are not normally independent. For example, it may be the case that if an object $x$ is of class $C$, $C(x)$, and it is related to another object $y$ through relation $R(x,y)$ then this other object $y$ should be in the same class $C(y)$. In logic: $\forall x \exists y ((C(x) \wedge R(x,y)) \imp C(y))$. Whether or not $C(y)$ holds will depend on the application: through reasoning, one may derive $C(y)$ where otherwise there might not have been evidence of $C(y)$ from training examples only; through learning, one may need to revise such a conclusion once  examples to the contrary become available. The vectorial representation proposed in this paper permits both reasoning and learning as exemplified above and detailed in the next section.

The above forms of reasoning and learning are integrated in a unifying 
framework, implemented within tensor networks, and exemplified in relational domains combining data and relational knowledge about the objects. It is expected that, through an adequate integration of numerical properties 
and relational knowledge, differently from the immediate related literature \cite{DBLP:journals/dagstuhl-reports/GarcezGHL14,AAAISpring,COCONIPS}, the framework introduced in this paper will be capable of combining in an effective way first-order logical inference on open domains with efficient relational multi-class learning using tensor networks.

The main contribution of this paper is two-fold. It introduces a novel framework for the integration of learning and reasoning which can take advantage of the representational power of  (multi-valued) first-order logic, and it instantiates the framework using tensor networks into an efficient implementation which shows that the proposed vector-based representation of the logic offers an adequate mapping between symbols and their real-world manifestations, which is appropriate for both rich inference and learning from examples.

The paper is organized as follows. In Section 2, we define Real Logic. In Section 3, we propose the Learning-as-Inference framework. In Section 4, we instantiate the framework by showing how Real Logic can be implemented in deep Tensor Neural Networks leading to Logic Tensor Networks (LTN). Section 5 contains an example of how LTN handles knowledge completion using (possibly inconsistent) data and knowledge from the well-known \emph{smokers and friends} experiment. Section 6 concludes the paper and discusses directions for future work. 

 
\def\headof{\mathsf{headOf}}
\def\arity{\alpha}
\def\bu{\mathbf{u}}
\def\bv{\mathbf{v}}
\section{Real Logic}
We start from a first order language $\L$, whose signature contains a
set $\C$ of constant symbols, a set $\Fn$ of functional symbols, and a
set $\Pr$ of predicate symbols.  The sentences of $\L$ are used to
express relational knowledge, e.g. the atomic formula $R(o_1,o_2)$
states that objects $o_1$ and $o_2$ are related to each other through 
binary relation $R$; $\forall xy.(R(x,y)\imp R(y,x))$ states that $R$
is a symmetric relation, where $x$ and $y$ are variables; $\exists y. R(o_1,y)$ 
states that there is an (unknown) object which is related to object $o_1$ 
through $R$. For simplicity, without loss of generality, we assume that all 
logical sentences of $\L$ are in prenex 
conjunctive, skolemised normal form \cite{Huth:2004:LCS:975331}, e.g. a sentence $\forall x (A(x)\imp
\exists y R(x,y))$ is transformed into an equivalent clause $\neg A(x)\vee R(x,f(x))$, where $f$
is a new function symbol. 

As for the semantics of $\L$, we deviate from the standard abstract
semantics of FOL, and we propose a \emph{concrete} semantics with
sentences interpreted as tuples of real numbers. To emphasise the fact
that $\L$ is interpreted in a ``real'' world, we use the term
\emph{(semantic) grounding}, denoted by $\G$, instead of the more
standard \emph{interpretation}%
\footnote{In logic, the term ``grounding'' indicates the
  operation of replacing the variables of a term/formula with
  constants. To avoid 
  confusion, we use the term ``instantiation'' for this.}.
 
\begin{itemize}
\item 
  $\G$ associates an $n$-tuple of real numbers $\G(t)$ to any closed
  term $t$ of $\L$; intuitively $\G(t)$ is the set of numeric features
  of the object denoted by $t$.
\item $\G$ associates a real number in the interval $[0,1]$ to each
  clause $\phi$ of $\L$. Intuitively, $\G(\phi)$ represents one's
  confidence in the truth of $\phi$; the higher the value, the higher
  the confidence. 
\end{itemize}
A grounding is specified only for the elements of the signature of $\L$. 
The grounding of terms and clauses is defined inductively, as follows. 
\begin{definition} 
A \emph{grounding} $\G$ for a first order language $\L$ is a function
from the signature of $\L$ to the real numbers that satisfies the following conditions: 
\begin{enumerate}
\item $\G(c)\in\R^n$ for every constant symbol $c\in\C$;
\item $\G(f)\in \R^{n\cdot \arity(f)}\longrightarrow \R^n$ for every $f\in\Fn$;
\item $\G(P)\in \R^{n\cdot \arity(R)}\longrightarrow [0,1]$ for every $P\in\Pr$;
\end{enumerate}
\end{definition}
A grounding $\G$ is inductively extended to all the closed terms and
clauses, as follows:
\begin{align*} 
\G(f(t_1,\dots,t_m)) & = \G(f)(\G(t_1),\dots,\G(t_m)) \\
\G(P(t_1,\dots,t_m))& = \G(P)(\G(t_1),\dots,\G(t_m)) \\ 
\G(\neg P(t_1,\dots,t_m)) & = 1-\G(P(t_1,\dots,t_m)) \\ 
\G(\phi_1\vee\dots\vee\phi_k) & = \mu(\G(\phi_1),\dots,\G(\phi_k))
\end{align*}
where $\mu$ is an s-norm operator, also known as a t-co-norm operator (i.e. the dual of some t-norm operator).
\footnote{Examples of t-norms which can be chosen here are 
Lukasiewicz, product, and  G\"odel. 
Lukasiewicz s-norm is defined as $\mu_{Luk}(x,y)=\min(x+y,1)$;
Product s-norm is defined as $\mu_{Pr}(x,y)=x+y-x\cdot y$;
G\"odel s-norm is defined as $\mu_{max}(x,y)=\max(x,y).$}

\begin{example}
  Suppose that $O=\{o_1,o_2,o_3\}$ is a set of documents defined on a finite
  dictionary $D=\{w_1,...,w_n\}$ of $n$ words. Let $\L$ be the language that
  contains the binary function symbol $concat(x,y)$ denoting the document resulting from the 
  concatenation of documents $x$ with $y$. Let $\L$ contain also the binary predicate $Sim$ which is supposed to be \emph{true} if document $x$ is deemed to be similar to document $y$.  
  An example of grounding is the one that associates to each
  document its bag-of-words vector \cite{Blei:2003:LDA:944919.944937}. As a consequence, a natural 
  grounding of the \emph{concat} function would be the sum of the 
  vectors, and of the \emph{Sim} predicate, the 
  cosine similarity between the vectors. More formally: 
  \begin{itemize}
    \item $\G(o_i) = \langle
      n^{o_i}_{w_1},\dots,n^{o_i}_{w_{n}}\rangle$, 
      where $n^d_w$ is the number of occurrences of word $w$ in document $d$;
    \item if $\bv,\bu\in\R^n$, $\G(concat)(\bu,\bv) = \bu+\bv$;
    \item if $\bv,\bu\in\R^n$, $\G(Sim)(\bu,\bv) = {\bu\cdot\bv \over
        ||\bu||||\bv||}$. 
  \end{itemize}
  For instance, if the three documents are $o_1$ = \textit{``John studies logic and plays football''}, $o_2$ = \textit{``Mary plays football and logic games''},  $o_3$ = \textit{``John and Mary play football and study logic together''}, and 
$W$=\{\textit{John, Mary, and, football, game, logic, play,
     study, together}\} then the following are 
   examples of the grounding of terms,
   atomic formulas and clauses. 
   \begin{align*}
     \G(o_1) & = \left<1,0,1,1,0,1,1,1,0\right> \\
     \G(o_2) & = \left<0,1,1,1,1,1,1,0,0\right> \\
     \G(o_3) & = \left<1,1,2,1,0,1,1,1,1\right> \\
     \G(concat(o_1,o_2)) & = \G(o_1) + \G(o_2) = \left<1,1,2,2,1,2,2,1,0\right> \\ 
     \G(Sim(concat(o_1,o_2),o_3) & = {\G(concat(o_1,o_2))\cdot\G(o_3)
                                   \over ||\G(concat(o_1,o_2))||\cdot||\G(o_3)||} 
       \approx {13 \over 14.83} \approx 0.88 \\ 
    \G(Sim(o_1,o_3)\vee Sim(o_2,o_3)) 
      & = \mu_{max}(\G(Sim(o_1,o_3),\G(Sim(o_2,o_3)) \\ 
      & \approx \max(0.86,0.73) = 0.86
   \end{align*}
 \end{example}
\vspace*{-1cm}

\section{Learning as approximate satisfiability} 
\def\pG{\hat{\mathcal{G}}}
\def\GT{\langle\K,\pG\rangle}
\def\vwphi{\langle[v,w],\phi\rangle}
\def\vwphiarg#1{\langle[v,w],\phi(#1)\rangle}
\def\vwphix{\vwphiarg{\bx}}
\def\vwphit{\vwphiarg{\bt}}
\def\bt{\mathbf{t}}
\def\bx{\mathbf{x}}
\def\Loss{\text{Loss}}
\def\parg{\theta}
\def\F{\mathcal{F}}
\def\AI{\mathit{AI}}
\def\GG{\mathbb{G}}

We start by defining ground theory and their satisfiability. 

\begin{definition}[Satisfiability]
Let $\phi$ be a closed clause in $\L$,
$\G$ a grounding, and $v\leq w\in[0,1]$. We say that $\G$ satisfies
$\phi$ in the confidence interval $[v,w]$, written 
$\G\models_v^w\phi$, if $v\leq\G(\phi)\leq w$. 
\end{definition}

A partial grounding, denoted by $\pG$, is a grounding that is defined
on a subset of the signature of $\L$. A grounded theory is a set of clauses 
in the language of $\L$ and partial
grounding $\pG$. 

\begin{definition}[Grounded Theory]
A \emph{grounded theory (GT)} is a pair $\GT$
where $\K$ is a set of pairs $\vwphix$, where 
$\phi(\bx)$ is a clause of $\L$ containing the set $\bx$ of free
variables, and $[v,w]\subseteq[0,1]$ is an interval 
contained in $[0,1]$, and $\pG$ is a partial grounding. 
\end{definition}



\begin{definition}[Satisfiability of a Grounded Theory]
  A GT $\GT$ is satisfiabile if there exists a grounding $\G$, 
  which extends $\pG$ such that for all
  $\vwphix\in\K$ and any tuple $\bt$ of closed terms,
  $\G\models_v^w\phi(\bt)$. 
\end{definition}






From the previous definiiton it follows that checking if a GT $\GT$ is
satisfiable amounts to seaching for an extension of the partial
grounding $\pG$ in the space of \emph{all possible groundings}, such
that \emph{all} the instantiations of the clauses in $\K$ are
satisfied w.r.t. the specified interval. Clearly this is unfeasible
from a practical point of view. As is usual, we must
restrict both the space of grounding and clause
instantiations. Let us consider each in turn:
 To check satisfiability on a subset of all the functions on
  real numbers, recall that a grounding should capture a latent correlation between the quantitative attributes of an object and its relational properties\footnote{For example, whether a document is classified as from the field of Artificial Intelligence (AI) depends on its bag-of-words grounding. If the language $\L$ contains the unary predicate $AI(x)$ standing for ``$x$ is a paper about AI'' then the grounding of $AI(x)$, which is a function from bag-of-words vectors to [0,1], should assign values close to $1$ to the vectors which are close semantically to $AI$. Furthermore, if two vectors are similar (e.g. according to the cosine similarity measure) then their grounding should be similar.}. In particular, we are interested in searching within a specific class of functions, in this paper based on tensor networks, although other family of functions can be considered. To limit the number of clause instantiations, which in general might be infinite since $\L$ admits function symbols, the usual approach is to consider the instantiations of each clause up to a certain depth \cite{DBLP:series/faia/Achlioptas09}.


When a grounded theory $\GT$ is inconsitent, that is, there is no
grounding $\G$ that satisfies it, we are interested in finding a grounding which satisfies \emph{as much as possible} of $\GT$. 
For any $\vwphi\in\K$ we want to find a grounding $\G$ that
  minimizes the \emph{satisfiability error}. An error occurs when a
  grounding $\G$ assigns a value $\G(\phi)$ to a clause $\phi$ which
  is outside the interval $[v,w]$ prescribed by $\K$. The measure of this error can be defined as the
  minimal distance between the points in the interval $[v,w]$ and
  $\G(\phi)$: 
  \begin{align}
  \label{eq:loss}
  \Loss(\G,\vwphi)=|x-\G(\phi)|, v\leq x\leq w
  \end{align}
  Notice that if $\G(\phi)\in[v,w]$, $\Loss(\G,\phi)=0$. 

The above gives rise to the following definition of
approximate satisfiability w.r.t. a family $\GG$ of grounding
functions on the language $\L$.

\begin{definition}[Approximate satisfiability]
  Let $\GT$ be a grounded theory and $\K_0$ a finite subset of the instantiations of the
  clauses in $\K$, i.e.
  $$K_0\subseteq\{\vwphit\}\mid\vwphix\in\K \mbox{ and $\bt$ is any $n$-tuple
    of closed terms.}\}$$
  Let $\GG$ be a family of grounding functions. We define the best
  satisfiability problem as the problem of finding an
  extensions $\G^*$ of $\pG$ in $\GG$ that minimizes the satisfiability error
  on the set $\K_0$, that is:
  $$
  \G^* = \argmin_{\pG\subseteq\G\in\GG} \sum_{\vwphit\in\K_0}\Loss(\G,\vwphit)
  $$
\end{definition}




\section{Implementing Real Logic in Tensor Networks} 
Specific instances of Real Logic can be obtained by selectiong the
space $\GG$ of groundings and the specific s-norm for the
interpretation of disjunction. In this section, we describe a 
realization of real logic where $\GG$ is the space of real
tensor transformations of order $k$ (where $k$ is a parameter). In
this space, function symbols are interpreted as linear
transformations. More precisely, if $f$ is a function symbol of arity 
$m$ and $\bv_1,\dots,\bv_m\in\R^n$ are real vectors
corresponding to the grounding of $m$ terms then 
$\G(f)(\bv_1,\dots,\bv_m)$ can be written as: 
$$
\G(f)(\bv_1,\dots,\bv_m)= M_f\bv + N_f
$$
for some $n\times mn$ matrix $M_f$ and $n$-vector $N_f$, 
where $\bv=\left<\bv_1,\dots,\bv_n\right>$.

The grounding of $m$-ary predicate $P$, $\G(P)$, is defined
as a generalization of the neural tensor network \cite{SocherChenManningNg2013} 
(which has been 
shown effective at knowledge compilation in the presence of simple 
logical constraints), as a 
function from $\R^{mn}$ to $[0,1]$, as follows:
\begin{align}
\label{eq:g-of-p}  
\G(P) = \sigma\left(u^T_P\tanh\left(\bv^TW_P^{[1:k]}\bv + V_P\bv + B_P\right)\right)
\end{align}
where $W_P^{[1:k]}$ is a 3-D tensor in $\R^{mn\times mn\times k}$,
$V_P$ is a matrix in $\R^{k\times mn}$, and $B_P$ is a vector in
$\R^{k}$, and $\sigma$ is the sigmoid function.  With this encoding,
the grounding (i.e. truth-value) of a clause can be determined by a
neural network which first computes the grounding of the literals
contained in the clause, and then combines them using the specific
s-norm. An example of tensor network for $\neg P(x,y)\imp A(y)$ is
shown in Figure~\ref{fig:example-of-tensor-net}.
\begin{figure}
\begin{center}
\begin{tikzpicture}[scale=.8]\tiny 
\draw[dashed,fill=green!10] (-.5,0) rectangle (5.5,5.8);
\draw[dashed,fill=violet!10] (6.5,0) rectangle (12.5,5.8);
\node at (0,5.5) {$\G(\neg P)$};
\node at (12,5.5) {$\G(A)$};
\node[rectangle,draw] (v) at (3,-.8) {$\bv = \left<v_1,\dots,v_n\right>$};
\node[rectangle,draw] (u) at (6,-.8) {$\bu = \left<u_1,\dots,u_n\right>$};
\node[circle,fill=black!10,draw] (WP1) at (0,1) {$W^1_P$};
\node[circle,fill=black!10,draw] (WP2) at (1,1) {$W^2_P$};
\node[circle,fill=black!10,draw] (VP1) at (2,1) {$V^1_P$};
\node[circle,fill=black!10,draw] (VP2) at (3,1) {$V^2_P$};
\node[circle,fill=black!10,draw] (BP1) at (4,1) {$B^1_P$};
\node[circle,fill=black!10,draw] (BP2) at (5,1) {$B^2_P$};
\node[circle,fill=white,draw] (plusP1) at (1.5,2.3) {$+$};
\node[circle,fill=white,draw] (plusP2) at (3.5,2.3) {$+$};
\node[circle,fill=white,draw] (tanhP1) at (1.5,3.1) {$th$};
\node[circle,fill=white,draw] (tanhP2) at (3.5,3.1) {$th$};
\node[circle,fill=black!10,draw] (UP) at (2.5,4) {$u_P$};
\node[circle,fill=white,draw] (sP) at (2.5,5) {$1-\sigma$};
\draw (v) -- (WP1);
\draw (v) -- (WP2);
\draw (v) -- (VP1);
\draw (v) -- (VP2);
\draw (u) -- (WP1);
\draw (u) -- (WP2);
\draw (u) -- (VP1);
\draw (u) -- (VP2);
\draw (WP1) -- (plusP1); 
\draw (WP2) -- (plusP2); 
\draw (VP1) -- (plusP1);
\draw (VP2) -- (plusP2);
\draw (BP1) -- (plusP1);
\draw (BP2) -- (plusP2);
\draw (plusP1) -- (tanhP1);
\draw (plusP2) -- (tanhP2);
\draw (tanhP1) -- (UP);
\draw (tanhP2) -- (UP);
\draw (UP) -- (sP);

\node[circle,fill=black!10,draw] (WA1) at (7+0,1) {$W^1_A$};
\node[circle,fill=black!10,draw] (WA2) at (7+1,1) {$W^2_A$};
\node[circle,fill=black!10,draw] (VA1) at (7+2,1) {$V^1_A$};
\node[circle,fill=black!10,draw] (VA2) at (7+3,1) {$V^2_A$};
\node[circle,fill=black!10,draw] (BA1) at (7+4,1) {$B^1_A$};
\node[circle,fill=black!10,draw] (BA2) at (7+5,1) {$B^2_A$};
\node[circle,fill=white,draw] (plusA1) at (7+1.5,2.3) {$+$};
\node[circle,fill=white,draw] (plusA2) at (7+3.5,2.3) {$+$};
\node[circle,fill=white,draw] (tanhA1) at (7+1.5,3.1) {$th$};
\node[circle,fill=white,draw] (tanhA2) at (7+3.5,3.1) {$th$};
\node[circle,fill=black!10,draw] (UA) at (7+2.5,4) {$u_A$};
\node[circle,fill=white,draw] (sA) at (7+2.5,5) {$\sigma$};
\draw (u) -- (WA1);
\draw (u) -- (WA2);
\draw (u) -- (VA1);
\draw (u) -- (VA2);
\draw (WA1) -- (plusA1); 
\draw (WA2) -- (plusA2); 
\draw (VA1) -- (plusA1);
\draw (VA2) -- (plusA2);
\draw (BA1) -- (plusA1);
\draw (BA2) -- (plusA2);
\draw (plusA1) -- (tanhA1);
\draw (plusA2) -- (tanhA2);
\draw (tanhA1) -- (UA);
\draw (tanhA2) -- (UA);
\draw (UA) -- (sA);
\node[circle,fill=white,draw] (max) at (6,6) {$max$};
\draw (sP) -- (max);
\draw (sA) -- (max);
\node[rectangle,draw] (output) at (6,7) {$\G(P(\bv,\bu)\imp A(\bu)$};
\draw (max) -- (output);
\end{tikzpicture}
\end{center}
\vspace{-.5cm}
\caption{Tensor net for $P(x,y)\imp A(y)$, with 
$\G(x)=\bv$ and $\G(y)=\bu$ and $k=2$.}
\label{fig:example-of-tensor-net}
\end{figure}
This architecture is a generalization of the
structure proposed in \cite{SocherChenManningNg2013}, that has been
shown rather effective for the task of knowledge compilation, also in 
presence of simple logical constraints. 
%
%
In the above tensor network 
formulation, $W_*, V_*, B_*$ and $u_*$ with $*\in\{P,A\}$ are parameters to be learned 
by  minimizing the loss function or, equivalently, to maximize 
the satisfiability of the clause $P(x,y)\imp A(y)$. 

\def\bl{\cellcolor{blue!20}}
\def\re{\cellcolor{red!20}}
\def\ye{\cellcolor{yellow!20}}
\section{An Example of Knowledge Completion}

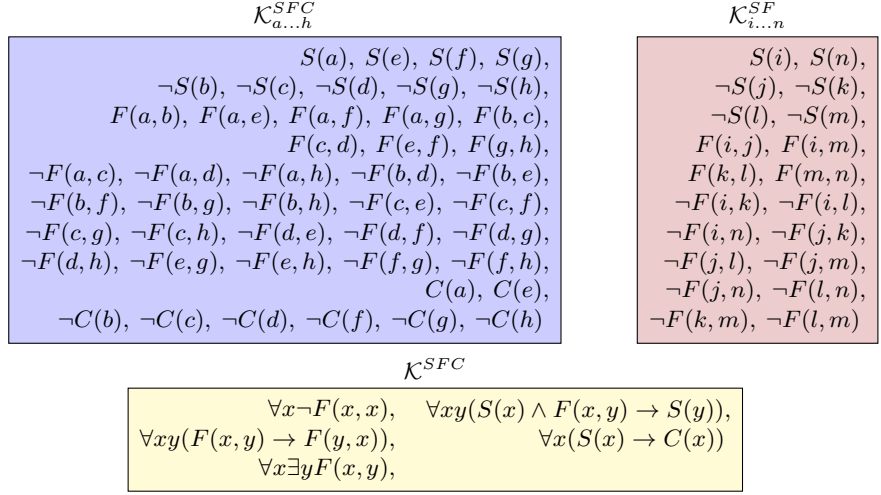
\begin{figure}
\begin{center}
\begin{tikzpicture}[y=27]\footnotesize
\node[label={$\K^{SFC}_{a\dots h}$},rectangle,draw,fill=blue!20] (kba-h) {
$\begin{array}{r}
S(a),\  
S(e),\  
S(f),\  
S(g),\ \\ 
\neg S(b),\
\neg S(c),\
\neg S(d),\
\neg S(g),\
\neg S(h),\ \\
F(a,b),\
F(a,e),\
F(a,f),\
F(a,g),\
F(b,c),\ \\
F(c,d),\
F(e,f),\
F(g,h),\ \\ 
\neg F(a,c),\
\neg F(a,d),\
\neg F(a,h),\
\neg F(b,d),\
\neg F(b,e),\ \\ 
\neg F(b,f),\ 
\neg F(b,g),\
\neg F(b,h),\
\neg F(c,e),\
\neg F(c,f),\ \\ 
\neg F(c,g),\
\neg F(c,h),\
\neg F(d,e),\
\neg F(d,f),\
\neg F(d,g),\ \\ 
\neg F(d,h),\
\neg F(e,g),\
\neg F(e,h),\
\neg F(f,g),\
\neg F(f,h),\ \\
C(a),\
C(e),\ \\
\neg C(b),\
\neg C(c),\
\neg C(d),\
\neg C(f),\
\neg C(g),\
\neg C(h)\ \ \\
\end{array}$};
\node[label={$\K^{SF}_{i\dots n}$}, rectangle,draw,right = of kba-h,fill=red!20] {
$\begin{array}{r}
S(i),\
S(n),\ \\ 
\neg S(j),\
\neg S(k),\ \\
\neg S(l),\
\neg S(m),\ \\
F(i,j),\
F(i,m),\ \\ 
F(k,l),\ 
F(m,n),\ \\ 
\neg F(i,k),\
\neg F(i,l),\ \\
\neg F(i,n),\
\neg F(j,k),\ \\
\neg F(j,l),\ 
\neg F(j,m),\ \\
\neg F(j,n),\ 
\neg F(l,n),\ \\
\neg F(k,m),\ 
\neg F(l,m)\ \
\end{array}$};
\node[label={$\K^{SFC}$}, rectangle,draw,fill=yellow!20] at (2,-3.5) {
$\begin{array}{r}
\forall x\neg F(x,x),\ \\
\forall xy(F(x,y)\imp F(y,x)),\ \\
\forall x\exists y F(x,y),\ \\ 
 \end{array}\ \ \ \ 
\begin{array}[b]{r}
 \forall xy(S(x) \wedge F(x,y) \imp S(y)),\ \\  
\forall x(S(x) \imp C(x))\ \ \\
\end{array}$};
\end{tikzpicture}
\end{center}
\vspace{-.3cm}
\caption{Knowledge-bases for the friends-and-smokers example.}
\label{fig:kbs}
\vspace{-.3cm}
\end{figure}

Logic Tensor Networks have been implemented as a 
Python library called \texttt{ltn} using Google's \tensorflow\ .
To test our idea, in this section we use the well-known 
\emph{friends and smokers\footnote{Normally, a probabilistic approach is taken to solve this problem, and one that requires instantiating all clauses to remove variables, essentially turning the problem into a propositional one; \texttt{ltn} takes a different approach.}} example \cite{Richardson-and-domingos-MLN-2006} to illustrate the task of knowledge completion in \texttt{ltn}.
There are 14 people divided into two groups $\{a,b,\dots,h\}$ and $\{i,j,\dots,n\}$. Within each group of people we have complete knowledge of their 
smoking habits. In the first group, we have complete knowledge of who has and  does not have cancer. In the second group, this is not known for any of the persons. Knowledge about the friendship relation is complete within each group only if symmetry of friendship is assumed. Otherwise, it is imcomplete in that it may be known that, e.g., $a$ is a friend of $b$, but not known whether $b$ is a friend of $a$. Finally, there is also general knowledge about smoking,
friendship and cancer, namely, that smoking causes cancer, friendship is normally a symmetric and anti-reflexive relation, everyone has a friend, and that smoking propagates (either actively or passively) among friends. All this knowledge can be  represented by the knowledge-bases shown in Figure~\ref{fig:kbs}. 

The facts contained in the knowledge-bases should have different degrees of
truth, and this is not known. Otherwise, the combined knowledge-base would be inconsistent (it would deduce e.g. $S(b)$ and $\neg S(b)$). Our main task 
is to complete the knowledge-base (KB), that is: (i) find the 
degree of truth of the facts contained in KB, (ii) find a truth-value for all 
the missing facts, e.g. $C(i)$, (iii) find the grounding of each
constant symbol $a,...,n.$\footnote{Notice how no grounding is
  provided about the signature of the knowledge-base.} To answer (i)-(iii), we use \texttt{ltn} to find a grounding that best approximates the complete KB. We start by assuming that all the facts contained in the knowledge-base are true (i.e. have degree of truth 1). To show the role of  background knolwedge in the learning-inference process, we run two experiments. In the first ($exp1$), we seek to complete a KB consisting of only factual knowledge: $\K_{exp1} = \K^{SFC}_{a\dots h} \cup \K^{SF}_{i\dots n}$. In the second ($exp1$), we also include background knowledge, that is:
$\K_{exp2}=\K_{exp1} \cup \K^{SFC}$. 

We confgure the network as follows: each constant (i.e. person) can have up to 30 
real-valued features. We set the number of layers $k$ in the tensor network to 10, 
and the regularization parameter%
\footnote{A smoothing factor $\lambda||\BOmega||^2_2$ is added to the loss function 
to create a preference for learned parameters with a lower absolute value.} $\lambda=1^{-10}$. 
For the purpose of illustration, we use the Lukasiewicz t-norm with s-norm 
$\mu(a,b) = \min(1,a+b)$, and use the harmonic mean as
aggregation operator. An estimation of the optimal grounding is 
obtained after 5,000 runs of the RMSProp learning algorithm
\cite{rmsprop-tieleman-hinton-2012} available in \tensorflow\ .

The results of the two experiments are reported in
Table~\ref{tab:experiments}. For readability, we use boldface for truth-values greater than 0.5. The truth-values of the facts
listed in a knowledge-base are highlighted with the same
background color of the knowledge-base in Figure \ref{fig:kbs}. The values with white background are the result of the knowledge completion produced by the LTN learning-inference procedure. To evaluate the quality of the results, one has to check whether 
(i) the truth-values of the facts listed in a KB are indeed 
close to 1.0, and (ii) the truth-values associated with knowledge completion correspond to expectation. An initial analysis shows that the LTN associated with $\K_{exp1}$ produces the same facts as $\K_{exp1}$ itself. In other words, the LTN fits the data. However, the LTN also learns to infer additional positive and negative facts about $F$ and $C$ not derivable from $\K_{exp1}$ by pure logical
reasoning; for example: $F(c,b)$, $F(g,b)$ and $\neg F(b,a)$. These
facts are derived by exploiting similarities between the groundings
of the constants generated by the LTN. For instance, $\G(c)$ and $\G(g)$ happen to present a high cosine similarity measure. As a result, facts about the friendship relations of $c$ affect the friendship relations of $g$ and vice-versa, for instance $F(c,b)$ and $F(g,b)$. 
The level of satisfiability associated with $\K_{exp1}\approx
1$, which indicates that $\K_{exp1}$ is classically
satisfiable. 

\begin{table}[t]\scriptsize\centering
\begin{tabular}{|c|}\hline
\ \ \ \ \ \ \ \ \begin{tabular}[t]{|l|l|l|l|l|l|l|l|l|l|l|} \hline
    &         &       & \multicolumn{8}{|c|}{$F$} \\  \cline{4-11}
    &  $S$ &  $C$  & $a$  & $b$  & $c$  & $d$  & $e$  & $f$  & $g$  & $h$  \\ \hline 
$a$ & \bl {\bf 1.00} & \bl {\bf 1.00} & 0.00 & \bl {\bf 1.00} & \bl 0.00 & \bl  0.00 & \bl  {\bf 1.00} &\bl{\bf 1.00} &\bl{\bf 1.00} &\bl0.00 \\
$b$ & \bl  0.00 & \bl  0.00 & 0.00 & 0.00 & \bl {\bf 1.00} &\bl 0.00 &\bl0.00 &\bl0.00 &\bl0.00 &\bl0.00 \\
$c$ & \bl  0.00 & \bl  0.00 & 0.00 & \bf 0.82 & 0.00 & \bl {\bf 1.00} &\bl0.00 &\bl0.00 &\bl0.00 &\bl0.00 \\
$d$ & \bl  0.00 & \bl  0.00 & 0.00 & 0.00 & 0.06 & 0.00 & \bl 0.00 & \bl 0.00 &\bl0.00 &\bl0.00 \\
$e$ & \bl {\bf 1.00} & \bl {\bf 1.00} & 0.00 & 0.33 & 0.21 & 0.00 & 0.00 & \bl {\bf 1.00} &\bl 0.00 &\bl 0.00 \\
$f$ &\bl {\bf 1.00} &\bl  0.00 & 0.00 & 0.00 & 0.05 & 0.00 & 0.00 & 0.00 & \bl 0.00 &\bl 0.00 \\
$g$ &\bl {\bf 1.00} &\bl  0.00 & 0.03 & {\bf 1.00} & {\bf 1.00} & {\bf 1.00} & 0.11 & {\bf 1.00} & 0.00 & \bl {\bf 1.00} \\
$h$ & \bl  0.00 & \bl 0.00 & 0.00 & 0.23 & 0.01 & 0.14 & 0.00 & 0.02 & 0.00 & 0.00 \\ \hline
\end{tabular}
\ \ \ \ 
\begin{tabular}[t]{|l|l|l|l|l|l|l|l|l|} \hline
       &     &       & \multicolumn{6}{|c|}{$F$} \\ \cline{4-9} 
       & $S$ &  $C$  & $i$  & $j$  & $k$  & $l$  & $m$  & $n$  \\ \hline 
$i$ &\re  {\bf 1.00} & 0.00 & 0.00 &\re {\bf 1.00} &\re 0.00 &\re 0.00 &\re {\bf 1.00} &\re 0.00 \\
$j$ &\re  0.00 &0.00 & 0.00 & 0.00 &\re 0.00 &\re 0.00 &\re 0.00 &\re 0.00 \\
$k$ &\re  0.00 &0.00 & 0.10 & {\bf 1.00} & 0.00 &\re {\bf 1.00} &\re 0.00 &\re 0.00 \\
$l$ &\re  0.00 &0.00 & 0.00 & 0.02 & 0.00 & 0.00 &\re 0.00 &\re 0.00 \\
$m$ &\re  0.00 &0.03 & {\bf 1.00} & {\bf 1.00} & 0.12 & {\bf 1.00} & 0.00 &\re {\bf 1.00} \\
$n$ &\re  {\bf 1.00} & 0.01 & 0.00 & 0.98 & 0.00 & 0.01 & 0.02 & 0.00 \\ \hline 
\end{tabular}\ \ \ \ \ \ \ \ 
\\ 
\normalsize Learning and reasoning on $\K_{exp1} = \K^{SFC}_{a\dots h} \cup
\K^{SF}_{i\dots n}$
\\ \\ 
\begin{tabular}[t]{|l|l|l|l|l|l|l|l|l|l|l|} \hline
       &     &       & \multicolumn{8}{|c|}{$F$} \\ \cline{4-11} 
       & $S$ &  $C$  & $a$  & $b$  & $c$  & $d$  & $e$  & $f$  & $g$  & $h$  \\ \hline 
$a$ &\bl \bf 0.84 &\bl \bf 0.87 & 0.02 &\bl \bf 0.95 &\bl 0.01 &\bl 0.03 &\bl \bf 0.93 &\bl \bf 0.97 &\bl \bf 0.98 &\bl 0.01 \\
$b$ &\bl 0.13 &\bl 0.16 & 0.45 & 0.01 &\bl \bf 0.97 &\bl 0.04 &\bl 0.02 &\bl 0.03 &\bl 0.06 &\bl 0.03  \\
$c$ &\bl 0.13 &\bl 0.15 & 0.02 & \bf 0.94 & 0.11 &\bl \bf 0.99 &\bl 0.03 &\bl 0.16 &\bl 0.15 &\bl 0.15  \\
$d$ &\bl 0.14 &\bl 0.15 & 0.01 & 0.06 & \bf 0.88 & 0.08 &\bl 0.01 &\bl 0.03 &\bl 0.07 &\bl 0.02  \\
$e$ &\bl \bf 0.84 &\bl \bf 0.85 & 0.32 & 0.06 & 0.05 & 0.03 & 0.04 &\bl \bf 0.97 &\bl 0.07 &\bl 0.06  \\
$f$ &\bl \bf 0.81 &\bl 0.19 & 0.34 & 0.11 & 0.08 & 0.04 & 0.42 & 0.08 &\bl 0.06 &\bl 0.05  \\
$g$ &\bl \bf 0.82 &\bl 0.19 & \bf 0.81 & 0.26 & 0.19 & 0.30 & 0.06 & 0.28 & 0.00 &\bl \bf 0.94  \\
$h$ &\bl 0.14 &\bl 0.17 & 0.05 & 0.25 & 0.26 & 0.16 & 0.20 & 0.14 & \bf 0.72 & 0.01  \\ \hline 
\end{tabular}
\ \ \ \ 
\begin{tabular}[t]{|l|l|l|l|l|l|l|l|l|} \hline
       &     &       & \multicolumn{6}{|c|}{$F$} \\ \cline{4-9} 
       & $S$ &  $C$  & $i$  & $j$  & $k$  & $l$  & $m$  & $n$  \\ \hline 
$i$ &\re  \bf 0.83 & \bf 0.86 &   0.02 &\re  \bf 0.91 &\re  0.01 &\re  0.03 &\re  \bf 0.97 &\re  0.01 \\
$j$ &\re  0.19 & 0.22 &   \bf 0.73 &  0.03 &\re  0.00 &\re  0.04 &\re  0.02 &\re  0.05 \\
$k$ &\re  0.14 & 0.34 &   0.17 &  0.07 &  0.04 &\re  \bf 0.97 &\re  0.04 &\re  0.02 \\
$l$ &\re  0.16 & 0.19 &   0.11 &  0.12 &  0.15 &  0.06 &\re  0.05 &\re  0.03 \\
$m$ &\re  0.14 & 0.17 &   \bf 0.96 &  0.07 &  0.02 &  0.11 &  0.00 &\re  \bf 0.92 \\
$n$ &\re \bf 0.84 & \bf 0.86 &   0.13 &  0.28 &  0.01 &  0.24 &  \bf 0.69 &  0.02 \\ \hline 
\end{tabular}
\\  \\ 
\begin{tabular}[t]{|l|c@{,}c|} \hline 
    & $a,\dots,h$ & $i,\dots,n$ \\ \hline 
$\forall x \neg F(x,x)$ &\multicolumn{2}{|c|}{\ye \bf 0.98} \\ 
$\forall xy(F(x,y)\imp F(y,x))$ &\ye \bf0.90 & \ye \bf 0.90 \\ 
$\forall x(S(x)\imp C(x))$ & \multicolumn{2}{|c|}{\ye \bf 0.77} \\ 
$\forall x(S(x)\wedge F(x,y)\imp S(y))$ &\ye \bf 0.96 &\ye \bf 0.92 \\ 
$\forall x\exists y (F(x,y))$ & \multicolumn{2}{|c|}{\ye \bf 1.0} \\ \hline 
\end{tabular}
\\ \\ 
\normalsize Learning and reasoning on $\K_{exp2}= \K^{SFC}_{a\dots h} \cup
\K^{SF}_{i\dots n}\cup \K^{SFC}$  \\ \\ \hline 
\end{tabular}
\normalsize \caption{\label{tab:experiments}}
\vspace{-1cm}
\end{table}

The results of the second experiment show that more facts can be learned with the inclusion of background knowledge. For example, the LTN now predicts that $C(i)$ and $C(n)$ are true. Similarly, from the symmetry of the friendship relation, the LTN concludes that $m$ is a friend of $i$, as expected. In fact, all the axioms in the generic background knowledge $\K^{SFC}$ are satisfied with a degree of satisfiability higher than 90\%, apart from the \emph{smoking causes cancer} axiom - which is responsible for the classical inconsistency since in the data $f$ and $g$ smoke and do not have cancer -, which has a degree of satisfiability of 77\%. 


\section{Related work}
\def\holds{\mathit{holds}} In his recent note,
\cite{towards-a-model-theory-for-distributed-representations-guha2015},
Guha advocates the need for a new model theory for distributed
representations (such as those based on embeddings). The note
sketches a proposal, where terms and (binary) predicates are all
interpreted as points/vectors in an $n$-dimensional real space. The
computation of the truth-value of the atomic formulae
$P(t_1,\dots,t_n)$ is obtained by comparing the projections of the
vector associated to each $t_i$ with that associated to $P_i$. Real
logic shares with 
\cite{towards-a-model-theory-for-distributed-representations-guha2015}
the idea that terms must be interpreted in a geometric space. It has, however, a different (and more general) interpretation of
functions and predicate symbols. Real logic is more general because the
semantics proposed in
\cite{towards-a-model-theory-for-distributed-representations-guha2015}
can be implemented within an \ltn\ with a single layer ($k=1$), since the
operation of projection and comparison necessary to compute the truth-value 
of $P(t_1,\dots,t_m)$ can be encoded within an
$nm\times nm$ matrix $W$ with the constraint that 
$\left<\G(t_1),\dots,\G(t_n)\right>^TW\left<\G(t_1),\dots,\G(t_n)\right>\leq
\delta$, which can be encoded easily in \ltn. 

Real logic is orthogonal to the approach taken by
(Hybrid) Markov Logic Networks (MLNs) and its variations 
\cite{Richardson-and-domingos-MLN-2006,DBLP:conf/aaai/WangD08,DBLP:conf/aaai/NathD15}. In
MLNs, the level of truth of a formula is determined by the number of
models that satisfy the formula: the more models, the higher the degree of truth. Hybrid MLNs introduce a dependency from the real features associated to constants, which is given, and not learned. In real logic, instead, the level of truth of a complex formula is determined by (fuzzy) logical reasoning, and the relations between the features of different objects is learned through error minimization. Another difference is that MLNs work under the \emph{closed world assumption}, while Real Logic is open domain. Much work has been done also on neuro-fuzzy approaches \cite{Kosko:1992:NNF:129386}. These are essentially propositional while real logic is first-order. 

Bayesian logic (BLOG) \cite{DBLP:conf/ijcai/MilchMRSOK05} is open domain, and in this respect  similar to real logic and LTNs. But, instead of taking an explicit probabilistic approach, LTNs draw from the efficient approach used by tensor networks for knowledge graphs, as already discussed. LTNs can have a probabilistic interpretation but this is not a requirement. Other statistical AI and probabilistic approaches such as lifted inference fall into this category, including probabilistic variations of inductive logic programming (ILP) \cite{DBLP:series/synthesis/2016Raedt}, which are normally restricted to Horn clauses. Metainterpretive ILP \cite{DBLP:journals/ml/MuggletonLT15}, together with BLOG, seem closer to LTNs in what concerns the knowledge representation language, but do not explore the benefits of tensor networks for computational efficiency. 
 
An approach for embedding logical knowledge onto data for the purpose of relational learning, similar to Real Logic, is presented in \cite{rocktaschel2015injecting}. Real Logic and \cite{rocktaschel2015injecting} share the idea of interpreting a logical alphabet in an $n$-dimensional real space. Terminologically, the term ``grounding'' in Real Logic corresponds to ``embeddings'' in \cite{rocktaschel2015injecting}. However, there are several differences. First,  \cite{rocktaschel2015injecting} uses function-free langauges, while we provide also groundings for functional symbols. Second, the model used to compute the truth-values of atomic formulas adopted in \cite{rocktaschel2015injecting} is a special case of the more general model proposed in this paper (as described in Eq.~\eqref{eq:g-of-p}). Finally, the semantics of the universal and existential quantifiers adopted in \cite{rocktaschel2015injecting} is based on the closed-world assumption (CWA), i.e. universally (respectively, existentially) quantified formulas are reduced to the finite conjunctions (respectively, disjunctions) of all of their possible instantiations; Real Logic does not make the CWA. Furthermore, Real Logic does not assume a specific t-norm. 

As in \cite{DBLP:journals/ml/DiligentiGMR12}, LTN is a framework for learning in the presence of logical constraints. LTNs share with \cite{DBLP:journals/ml/DiligentiGMR12} the idea that logical constraints and training examples can be treated uniformly as supervisions of a learning algorithm. LTN introduces two novelties: first, in LTN existential quantifiers are not grounded into a finite disjunction, but are \emph{scolemized}. In other words, CWA is not required, and  existentially quantified formulas can be satisfied by ``new individuals''. Second, LTN allows one to generate data for prediction. For instance, if a grounded theory contains the formula $\forall x \exists y R(x,y)$, LTN generates a real function (corresponding to the grounding of the Skolem function introduced by the formula) which for every vector $\bv$ returns the feature vector $f(\bv)$, which can be intuitively interpreted as being the set of features of a \emph{typical} object which takes part in relation $R$ with the object having features equal to $\bv$.

Finally, related work in the domain of neural-symbolic computing and neural network fibring \cite{DBLP:series/cogtech/GarcezLG2009} has sought to combine neural networks with ILP to gain efficiency \cite{DBLP:journals/ml/FrancaZG14} and other forms of knowledge representation, such as propositional modal logic and logic programming. The above are more tightly-coupled approaches. In contrast, LTNs use a richer FOL language, exploit the benefits of knowledge compilation and tensor networks within a more loosely- coupled approach, and might even offer an adequate representation of equality in logic. Experimental evaluations and comparison with other neural-symbolic approaches are desirable though, including the latest developments in the field, a good snapshot of which can be found in \cite{COCONIPS}.


\vspace*{-0.3cm}
\section{Conclusion and future work}
We have proposed \emph{Real Logic}: a uniform framework for
learning and reasoning. Approximate satisfiability is defined as a learning 
task with both knowledge and data being mapped onto real-valued vectors. 
With an inference-as-learning approach, relational knowledge constraints and state-of-the-art 
data-driven approaches can be integrated. We showed how real logic can be 
implemented in deep tensor networks, which we call Logic Tensor Networks (LTNs), and 
applied efficiently to knowledge completion and data prediction tasks. As future work, we 
will make the implementation of LTN available in \tensorflow\ and apply it to large-scale 
experiments and relational learning benchmarks for comparison with statistical relational learning, neural-symbolic computing, and (probabilistic) inductive logic programming approaches. 
\vspace*{-0.3cm}
\bibliographystyle{plain}
\small
\bibliography{biblio}
\end{document}